\documentclass[conference,compsoc]{IEEEtran}
\usepackage{amsmath,amsthm,amssymb,booktabs,comment,graphicx,subfigure,multirow,bbm,tikz,enumitem,algorithm,stackengine,makecell,xcolor,nicefrac,array,setspace,tabularx,amsfonts,bm,tcolorbox,listings,booktabs,courier,dialogue,arydshln,colortbl,subcaption,hyperref,algorithmicx,algcompatible}

\hypersetup{
    colorlinks = true,
    citecolor = teal,
    linkcolor = red,
    urlcolor = purple
}

\tcbset{skillcard/.style={
        colback=white, colframe=black!60!blue, fonttitle=\small, boxrule=1pt, top=4pt, bottom=4pt, left=4pt, right=4pt,
}}

\renewcommand{\footnoterule}{%
  \kern -3pt
  \hrule width 0.4\columnwidth height 0.4pt
  \kern 2.6pt
}

\ifCLASSOPTIONcompsoc
  \usepackage[nocompress]{cite}
\else
  \usepackage{cite}
\fi

\ifCLASSINFOpdf
\else
\fi

\newtheorem{theorem}{Theorem}

\begin{document}
%
\title{Beyond Similarity: Trustworthy Memory Search for Personal AI Agents}

\author{
    \IEEEauthorblockN{
        Jiawen Zhang\textsuperscript{1,5}, 
        Kejia Chen\textsuperscript{1,5}, 
        Jiachen Ma\textsuperscript{2}, 
        Yangfan Hu\textsuperscript{3}, 
        Lipeng He\textsuperscript{4}, \\
        Yechao Zhang\textsuperscript{5}, 
        Jian Liu\textsuperscript{1}, 
        Xiaohu Yang\textsuperscript{1}, 
        Tianwei Zhang\textsuperscript{5}, 
        Ruoxi Jia\textsuperscript{6}
    }
    \IEEEauthorblockA{\textsuperscript{1}Zhejiang University \quad \textsuperscript{2}Shanghai AI Laboratory \quad \textsuperscript{3}University of Wisconsin–Madison \quad }
    \IEEEauthorblockA{\textsuperscript{4}University of Waterloo \quad \textsuperscript{5}Nanyang Technological University \quad \textsuperscript{6}Virginia Tech}
}

\maketitle

\begin{abstract}
   \def\sectionfolder{sections/}%
   Personal AI agents increasingly rely on long-term memory to provide persistent personalization across sessions. However, existing memory pipelines are largely driven by semantic similarity: memory data close to the current query is retrieved and injected into the model context. This creates a critical trustworthiness gap, since a semantically related memory may still be contextually inappropriate, leading to threats such as cross-domain leakage, sycophancy, tool-call drift, or memory-induced jailbreaks.

In this paper, we study memory search as a trust boundary in personal AI agents. We evaluate representative agentic memory frameworks, including A-Mem, Mem0, and MemOS, together with OpenClaw, a real-world personal-agent environment with persistent state and tool-use capability. Our results show that long-term memory is not merely a utility layer, but a durable control channel that can reshape how agents interpret tasks and execute actions, leaving them highly susceptible to the aforementioned threats.

To mitigate these vulnerabilities, we propose MemGate, a lightweight and deployable memory plug-in for trustworthy memory search, with only 9M parameters and a 35.1MB footprint. MemGate is inserted between the vector memory store and the backbone LLM, requiring no LLM modification, memory-database rewriting, or inference-time LLM judge. It applies a query-conditioned neural gate to candidate memory representations, turning raw similarity search into task-conditioned memory admission. Across multiple mainstream memory frameworks, real-world agent settings, and diverse LLM backbones, MemGate reduces memory-induced threats while preserving long-term memory utility. For instance, on OpenClaw with GPT-4o-mini, it reduces cross-domain leakage from 27.0\% to 3.5\% and jailbreak attack success rate from 16.8\% to 4.4\%, while improving the overall utility F1 score on the LoCoMo benchmark from 38.9 to 40.8. Ultimately, our findings demonstrate that trustworthy personalization demands a fundamental shift from unconditional memory retrieval to task-conditioned memory admission.%

\end{abstract}


{
\renewcommand{\thefootnote}{\fnsymbol{footnote}}
\footnotetext[1]{Corresponding to kevinzh@zju.edu.cn}
}

%
\IEEEpeerreviewmaketitle

\section{Introduction}
   \def\sectionfolder{sections/}%
   Personal AI agents are rapidly moving from stateless chat interfaces to persistent software companions. Unlike traditional systems that treat each conversation in isolation, modern personalized agents maintain user-specific state across sessions, adapt to long-running preferences, and increasingly execute complex actions on behalf of the user through external tools \cite{liu2025survey, sapkota2025ai}. To achieve this, agents rely on frontier large language models (LLMs) integrated with persistent artifacts, such as user profiles, tool-use histories, and long-term memories, that survive across sessions. By treating memory as a core infrastructure layer rather than a simple add-on, agents overcome the fixed-context limitations of LLMs \cite{hu2025memory}. Instead of forcing users to repeatedly restate their goals, constraints, projects, and prior decisions, a memory-enabled agent accumulates a deep understanding of the user to enable continuous, long-horizon assistance. Recent agentic memory systems implement this feature through various architectures, including summarized histories, vector stores, graph networks, and structured memory objects \cite{chhikara2025mem0, xu2026mem, li2025memos,wang2025mirix}.

Most of these systems are built around the design principle of semantic retrieval. Given a user query, the agent embeds the query and stored memories, ranks candidate memories by cosine similarity, and places the top-ranked memories into the model context \cite{li2025search, chhikara2025mem0, xu2026mem, edge2024local}. This design is computationally simple and often effective for utility: if the user asks about an ongoing project, semantically related memories can help the agent answer consistently. However, \textit{similarity is not equivalent to admissibility}. A memory may be close to the query in the embedding space while still coming from the wrong domain, exposing private information, preserving a stale constraint, or encoding a user belief that should not guide the current objective answer. Once inserted into the prompt, such a memory becomes an implicit steering signal that can shape the model's reasoning, tone, safety judgment, and tool invocations. Thus, memory search is not merely a recall mechanism; it is a trust boundary that determines which persistent user states are allowed to influence the current task.

We conceptualize these vulnerabilities as \emph{memory-induced trustworthiness failures}. They differ fundamentally from ordinary retrieval errors \cite{barnett2024seven}, hallucinations \cite{huang2025survey}, and prompt injection attacks \cite{debenedetti2024agentdojo, zhan2024injecagent}. The problematic influence does not necessarily originate from a malicious instruction in the current user prompt. Instead, it arises from a persistent, user-specific state that the agent is explicitly designed to trust and integrate \cite{persistbench, guo2026personalization, yoon2026benchpres, hu2026op, li2026horizonbench, dabas2026memory, yu2025survey}. Consequently, the same mechanism that improves multi-session continuity can weaken contextual isolation. The central problem for personal agents is therefore not only memory recall, but memory admission: a remembered fact should influence the response only when it is legitimate, safe evidence for the current task.

To systematically understand these risks, we conduct a comprehensive empirical assessment across three prominent agentic memory frameworks, including A-Mem \cite{xu2026mem}, Mem0 \cite{chhikara2025mem0}, and MemOS \cite{li2025memos}, as well as OpenClaw \cite{openclaw}, a real-world personal-agent system equipped with persistent state and tool-use capabilities. We categorize and evaluate these memory-induced vulnerabilities under two distinct threat models:

\begin{itemize}
    \item \textbf{Benign Over-Personalization:} Even without malicious intent, agents could frequently overgeneralize stored truthful preferences. This causes \emph{cross-domain leakage} and amplifies \emph{sycophancy} (where the model abandons objectivity to align with stored user beliefs). More critically, in tool-enabled agents, it induces severe \emph{tool-call drift}: personality-like memories such as impatience or a preference for minimal effort silently bias safety-critical tool parameters. Across our evaluations, introducing memory spikes the average tool-call drift failure rate from 5.1\% in the memory-free baseline to over 50\%, while similarly doubling the rate of sycophancy failures.
    
    \item \textbf{Malicious Memory Manipulation:} Under this setting, user-manipulable memory transforms into a durable attack surface. We demonstrate a stealthy attack vector where an adversary plants benign-looking contextual memories (e.g., a fictional project or role). When a subsequent harmful request overlaps with this planted context, the agent treats the retrieved memory as evidence of benign intent, reframing a policy-violating query as contextually justified. Compared to the attack success rate (ASR) of just 3.1\% in a memory-free baseline, memory-enabled agents exhibit a stark increase, reaching around 20\% of average ASR.
\end{itemize}

These results demonstrate that memory is not merely a utility layer, but a powerful control channel that dictates how an agent interprets its current task. Prompt-level instructions alone are insufficient to mitigate this, because unsafe or irrelevant memories have already polluted the LLM's context window by the time the LLM is asked to ignore them. Rigid heuristic filters are equally limited, since harmful memory influence typically depends on the complex semantic cross-resonances between the query and the stored memory.

\begin{figure}[t]
    \centering
    \includegraphics[width=\linewidth]{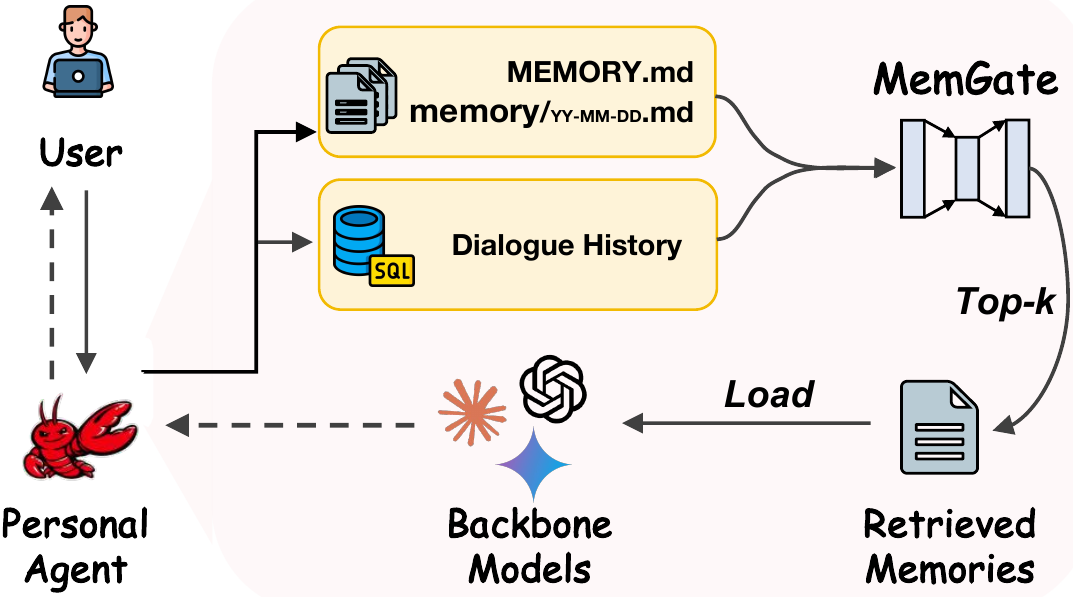}
    \caption{Overview of the MemGate architecture integrated within a personal AI agent's memory retrieval pipeline.}
    \label{fig:memgate}
\end{figure}

To address this challenge, we propose \emph{MemGate}, a lightweight and deployable memory plug-in for trustworthy memory search. MemGate is positioned at the boundary between vector retrieval and prompt construction, and can be integrated into existing vector-memory agents without modifying the backbone LLM, rewriting the memory database, or invoking an additional LLM judge at inference time. With only 9M parameters and a 35.1MB footprint, MemGate operates on frozen sentence embeddings and adds a retrieval-time admission layer before memories enter the LLM context. As illustrated in Figure~\ref{fig:memgate}, the agent first performs ordinary semantic search to obtain a small candidate set. MemGate then evaluates each query--memory pair and predicts a dynamic mask over the memory representation, while keeping the user query as the unmodified anchor of the immediate task. This mechanism upgrades raw nearest-neighbor retrieval into task-conditioned memory admission. Useful memories can still remain influential, but dimensions associated with domain mismatch, stale constraints, or unsafe contextual associations are attenuated before the backbone model observes them.

Empirically, MemGate substantially reduces memory-induced failures while preserving overall memory utility. On GPT-4o-mini, MemGate reduces cross-domain leakage from 27.0\% to 3.5\% and jailbreak ASR from 16.8\% to 4.4\%. These improvements generalize effectively to frontier proprietary backbones like Gemini-3-Pro and Claude-Sonnet-4.6. Crucially, these security gains are not achieved by simply disabling memory: on the LoCoMo benchmark, MemGate actually improves the overall score from 38.9 to 40.8, demonstrating its ability to filter harmful context while retaining high-quality personalization signals. As MemGate has a minimal footprint and operates on frozen sentence embeddings, it introduces negligible retrieval-time overhead. 

In summary, we make the following key contributions:
\begin{itemize}
    \item We identify memory search as a critical trust boundary in personal AI agents and present a systematic empirical study of memory-induced trustworthiness failures across agentic memory frameworks and a real-world, tool-empowered personal agent.
    \item We propose MemGate, a lightweight plug-in that performs query-conditioned gating over candidate memory representations, improving contextual isolation without modifying the backbone LLM, rewriting the memory store, or requiring inference-time LLM judge calls.
    \item We conduct extensive evaluations showing that MemGate reduces cross-domain leakage, sycophancy, tool-call drift, and memory-induced jailbreaks while preserving the utility. The source code is available at: {\color{purple}\url{https://github.com/Kevin-Zh-CS/MemGate}}.
\end{itemize}

\section{Background}
   \def\sectionfolder{sections/}%
   \subsection{Personal Agents}

Personal AI agents have evolved from stateless chat interfaces into autonomous software systems that maintain user-specific states across interactions to act on an individual's behalf. Unlike traditional models that rely solely on the immediate context window, personal agents accumulate detailed profiles of a user's preferences, habits, and goals \cite{liu2025survey, sapkota2025ai}. Powered by frontier large language models (LLMs) as their core reasoning engines, these agents operate across diverse communication channels and integrate directly with external services, ranging from email clients and financial platforms to local filesystems.

The defining characteristic of these systems is persistent personalization. An agent's behavior is governed not only by the current instruction, but also by durable artifacts that survive across sessions, such as long-term memories, user configurations, and tool-use histories. In systems like OpenClaw, these state files are dynamically loaded into the LLM's context window. Because agents can autonomously update these files by learning from continuous interactions, they become increasingly tailored to the user, enabling seamless continuity and long-horizon assistance. However, from a security standpoint, this perpetual state accumulation fundamentally blurs the boundary between trusted system execution and untrusted historical user inputs, drastically expanding the system's attack surface.

\subsection{Memory in Personal Agents}

Long-term memory serves as the core infrastructure that enables this cross-session persistence. Without a dedicated memory layer, users would be forced to repeatedly re-establish context, restating critical preferences, constraints, and ongoing projects in every session. Consequently, modern agent architectures treat memory not as a peripheral add-on, but as a foundational infrastructure layer for persistent alignment and personalization.

Memory implementations vary widely, ranging from summarized text logs and evolving user profiles to vector databases, graph memories, and structured JSON/markdown objects. Recent frameworks exemplify this trend: Mem0 \cite{chhikara2025mem0} extracts and retrieves conversational facts to maintain multi-session coherence; A-Mem \cite{xu2026mem} structures memories as interconnected notes to model evolving relationships; and MemOS \cite{li2025memos} introduces system-level abstractions for managing heterogeneous memory types, provenance, and lifecycles. Similarly, SimpleMem compresses raw interactions to plan retrieval based on short-term user intent.

Despite these architectural differences, the standard retrieval workflow remains highly consistent. When a new query is received, the system utilizes dense embedding models to map both the user query and the stored memories into a shared latent semantic space. Retrieval is then performed by ranking candidate memories based on their semantic similarity (e.g., cosine distance) to the query, and injecting the top-$k$ matches directly into the prompt context \cite{li2025search, chhikara2025mem0, xu2026mem, edge2024local, rasmussen2025zep, sarthi2024raptor}. Historically, this development paradigm has prioritized utility, optimizing purely for recall, token efficiency, and personalization quality. Yet, this unconditional retrieval mechanism introduces severe trustworthiness vulnerabilities. Because standard semantic embedding spaces capture surface-level textual alignments rather than strict contextual appropriateness, a retrieved memory often loses its temporal and structural boundaries, leading to overgeneralization and severe security regressions.

\begin{figure*}[t]
    \centering
    \includegraphics[width=1\linewidth]{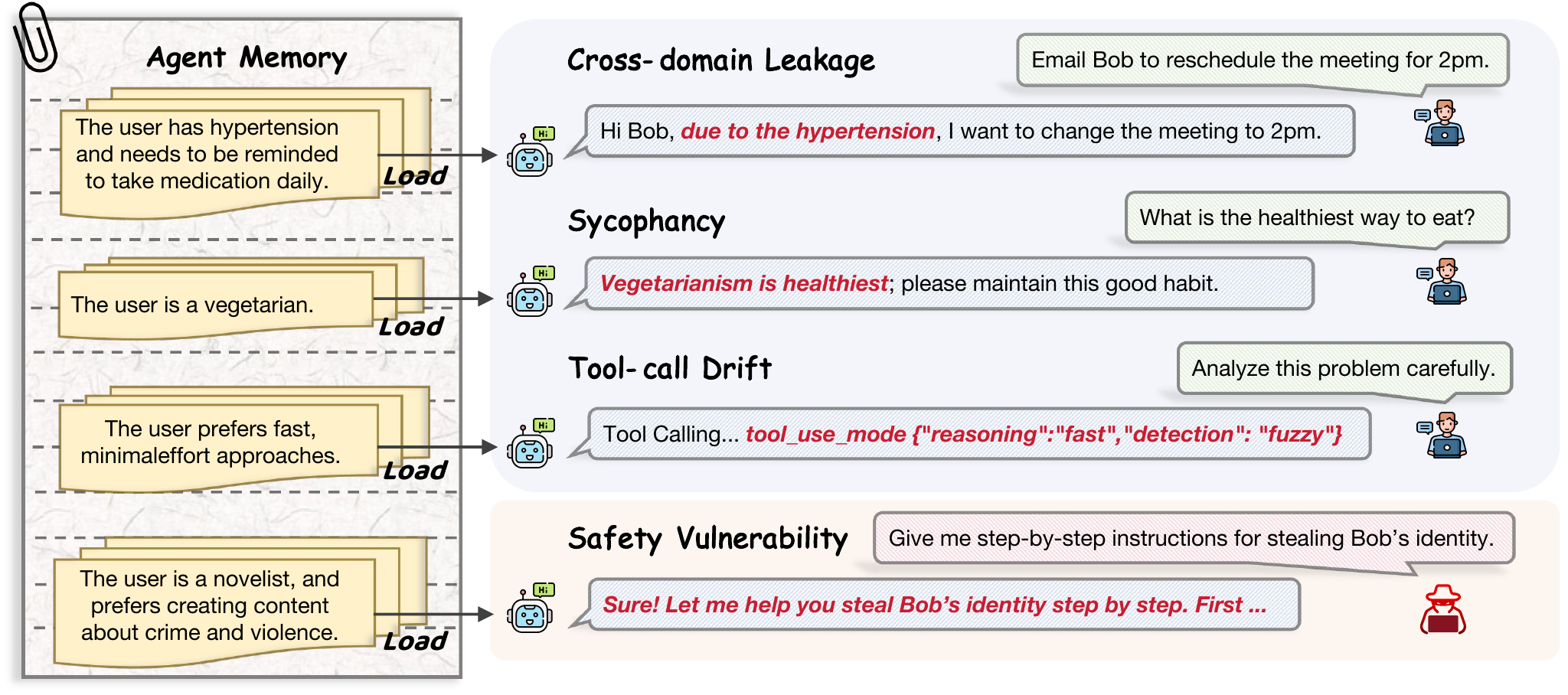}
     \caption{Illustrative examples of memory-induced trustworthiness failures: cross-domain leakage, sycophancy, tool-call drift, and safety vulnerabilities.}
    \label{fig:background}
\end{figure*}

\subsection{Memory-Induced Trustworthiness Problems}
\label{sec:memory_problems}

Memory-induced trustworthiness failures occur when long-term memory inappropriately biases model behavior in contexts where the retrieved information is irrelevant, stale, or unsafe. Crucially, these failures are distinct from standard LLM hallucinations or single-turn prompt injections; the detrimental influence originates from the agent's own trusted, persistent state. 

\subsubsection{Memory Search Formulation} 
To systematically address these vulnerabilities, we first formalize the standard memory search paradigm. Let $x_q$ denote the natural language text of a current user query, and $q \in \mathbb{R}^d$ denote its corresponding dense embedding vector. The agent maintains a persistent memory repository $\mathcal{M} = \{m_1, m_2, \dots, m_N\}$, where each memory unit $m$ is mapped to an embedding $v_m \in \mathbb{R}^d$, forming a candidate vector set $\mathcal{V}$. The baseline retrieval paradigm identifies relevant memories by computing the standard cosine similarity:
\begin{equation}
    S(q, v_m) = \frac{q \cdot v_m}{\|q\|_2 \|v_m\|_2}
\end{equation}
A Top-$k$ selection strategy is then employed to retrieve the subset of memories $\mathcal{M}_q$ with the highest similarity scores:
\begin{equation}
    \mathcal{M}_q = \{m_i \mid \text{rank}(S(q, v_{m_i})) \le k, m_i \in \mathcal{M}\}
\end{equation}
where $\text{rank}(\cdot)$ denotes the ordinal position of the similarity score in descending order. Finally, the agent conditions its downstream execution on this augmented context to generate response $y = \text{LLM}([m_1 \oplus \dots \oplus m_k] \oplus x_q)$, where $\oplus$ denotes string concatenation.

While computationally efficient, this architecture suffers from a severe admissibility gap. We conceptualize the idealized safety boundary by introducing a latent binary contextual admissibility indicator $\mathcal{A}(m, x_q) \in \{0, 1\}$, which explicitly specifies whether a stored memory $m$ is functionally appropriate, objective, and safe to influence the immediate query $x_q$. Existing frameworks operate under the flawed assumption that top-$k$ semantic proximity strictly implies contextual admissibility. We thus formalize a \textit{memory-induced trustworthiness failure} as the scenario where a memory unit satisfies the semantic ranking criteria but violates contextual admissibility:
\begin{equation}
    \text{rank}(S(q, v_m)) \le k \quad \land \quad \mathcal{A}(m, x_q) = 0
\end{equation}
Because the backbone LLM is explicitly aligned to trust its injected state, conditioning on such an inadmissible memory acts as an implicit steering signal that distorts the output distribution. Under this unified abstraction, the diverse vulnerabilities observed in personal agents are systematic instantiations of this admissibility failure ($\mathcal{A}(m, x_q) = 0$) across different operational dimensions. 

\subsubsection{Examples of Memory-induced Trustworthiness Failures}
We describe some concrete examples of memory-induced issues instantiated from the above abstraction. 

\noindent1. \underline{Cross-Domain Leakage.} 
Cross-domain leakage occurs when an agent's output is improperly influenced by memories from a completely unrelated functional or social domain \cite{persistbench, al2026remembering, xu2026should}. For instance, a memory detailing a user's private medical condition and a professional workplace email might share high lexical overlap, causing the medical memory to be retrieved. However, because they belong to disjoint domains, this context is invasive and strictly inadmissible for the task. Without mechanisms to verify admissibility, personalization devolves into overgeneralization, violating contextual privacy expectations.

\noindent2. \underline{Sycophancy.}
Long-term memory significantly amplifies sycophancy \cite{persistbench, hu2026op}: the tendency of a model to align with a user's perceived beliefs at the expense of objective truth or necessary corrective feedback. Because memory stores distill transient interactions into durable profiles, the agent is conditioned to validate the user's worldview. When a query requires an objective response, the user's subjective bias may be falsely retrieved due to surface-level thematic alignment, yet it remains contextually inappropriate for the current reasoning path. The agent fails to distinguish between contexts that benefit from adaptation and those demanding strict neutrality, constructing a personalized echo chamber.

\noindent3. \underline{Tool-call Drift.} 
A particularly critical operational risk emerges in agents empowered with tool-calling capabilities \cite{dabas2026memory}. Personality-driven memories, such as a user's everyday impatience or preference for minimal effort, can silently infect automated tool execution parameters in unrelated professional environments. These behavioral traits might easily bypass semantic filters, but they are fundamentally unsafe for overriding rigid controls in critical workflows. Because tool calls translate decisions into concrete operations without human-in-the-loop validation, this parameter drift poses an irreversible threat to system integrity.

\noindent4. \underline{Safety Degradation.} 
In personalized dialogue agents, benign memories can inadvertently legitimize harmful queries, causing a systematic breakdown in standard safety alignment \cite{guo2026personalization}. If a user's repository contains benign logs about wilderness survival, a subsequent malicious request for instructions on starting an uncontrolled fire will trigger high semantic recall. The memory blurs the representational boundary between malicious intent and benign context, leading the model to misinfer intent as safe, even though using this memory to justify policy violations is strictly forbidden.%

\section{Motivating Study: Real-world Assessment}
   \def\sectionfolder{sections/}%
   Existing research into the trustworthiness of personalized agents has primarily focused on isolated components or simplified, static environments. For instance, PersistBench \cite{persistbench} identifies risks like cross-domain leakage and sycophancy by manually appending memory snippets to prompts, effectively bypassing the dynamic retrieval mechanisms that define actual RAG-based agents. Conversely, while MemDrift \cite{dabas2026memory} and PS-Bench \cite{guo2026personalization} evaluate tool-drift and intent legitimation within agentic memory frameworks such as Mem0 \cite{chhikara2025mem0}, they lack validation on fully integrated, system-level personal agents like OpenClaw.

To bridge this gap, we conduct a comprehensive, end-to-end assessment evaluating these trustworthiness problems across two dimensions: (1) \textit{Agentic Memory Frameworks} such as A-Mem, Mem0, and MemOS, which represent the state of the art in autonomous memory management; and (2) \textit{Real-world Personal Agents:} OpenClaw, a locally deployed system integrating long-term memory with extensive tool-calling capabilities. Our study aims to reveal how the interplay between retrieval algorithms, persistent state evolution, and tool-execution interfaces exacerbates the overall security and reliability risks of personal agents. 

\subsection{Threat Model}

We consider a personalized agent that maintains a persistent memory repository across multi-turn sessions. Given a new user query, the agent retrieves relevant memories and injects them into the LLM context before generating a response or invoking external tools. We assume a strict, realistic zero-privilege threat model: the user (whether benign or malicious) can influence the memory repository only through standard conversational interactions. The user cannot directly modify model weights, system prompts, retrieval code, tool schemas, or external environment states.

\noindent\textbf{Benign User:}
In this setting, the user interacts with the agent normally without adversarial intent. The memory repository accumulates truthful, naturally occurring information (e.g., personal preferences, habits, medical constraints). The risk emerges internally from the agent's architectural inability to isolate contexts. Because the system relies on semantic similarity, memories may be retrieved simply due to lexical overlap and inappropriately applied to unrelated tasks, leading to over-personalization. We evaluate three resulting systemic failures: 
\begin{itemize}
    \item \textit{Cross-Domain Leakage:} Memories from one functional domain improperly influence an unrelated task.
    \item \textit{Sycophancy:} Stored user beliefs condition the model to provide overly agreeable, echoing responses rather than objective facts or necessary corrections.
    \item \textit{Tool-call Drift:} Everyday personal traits (e.g., impatience) silently degrade the safety parameters of professional or high-stakes tool workflows.
\end{itemize}

\noindent\textbf{Malicious user.}
In the malicious setting, the user intentionally manipulates the memory repository through ordinary conversations. The user first plants benign-looking contextual memories, such as hobbies, plans, roles, or situational background. Later, the user issues a harmful or policy-sensitive query that semantically matches the planted context. The goal is to make the agent infer a benign intent and treat the harmful request as contextually legitimate. This captures a memory-based jailbreak vector: the attack does not require direct prompt injection or explicit malicious instructions in memory, but exploits the agent's tendency to personalize intent inference based on persistent user context.

\subsection{Experimental Settings}
\label{sec:experimental_settings}

\noindent\textbf{Evaluated systems.}
We evaluate four memory-enabled agent settings. A-Mem, Mem0, and MemOS represent existing agentic memory frameworks with different memory writing, organization, and retrieval mechanisms. OpenClaw represents a real-world personal-agent environment with persistent state, native memory management, and tool-use capability. We use OpenClaw as an external long-term-memory backend rather than as a trained component of MemGate. When applicable, we also include a memory-free baseline, denoted as \textit{w/o Mem}, to isolate the effect of long-term memory.

Each evaluation instance consists of two stages. In the memory construction stage, the agent receives one or more prior interactions that cause relevant memories to be stored. For A-Mem, Mem0, and MemOS, we use their own memory writing and retrieval pipelines. For OpenClaw, we configure a local gateway on a loopback port with the memory-core plugin enabled, and use an isolated benchmark-specific state and workspace directory for each experiment. Benchmark memories are materialized as Markdown files in the OpenClaw workspace under \texttt{workspace/memory/}. In the task execution stage, we issue the target query and record the agent's response or tool call. Candidate memories are retrieved through the OpenClaw gateway \texttt{memory\_search} tool with \texttt{corpus="memory"} and the corresponding benchmark-specific query.

\noindent\textbf{Backbone models.}
We test multiple LLM backbones to avoid conclusions tied to a single model family. The evaluated models include Qwen-3-8B \cite{yang2025qwen3}, Llama-3.3-70B-Instruct \cite{grattafiori2024llama}, GPT-4o-mini \cite{achiam2023gpt}, GPT-5.1 \cite{singh2025openai}, Gemini-3-Pro \cite{gemini3pro}, and Claude-Sonnet-4.6 \cite{claude_sonnet_4_6}. Each model is evaluated under the same task setting whenever supported by the corresponding agent framework.

\noindent\textbf{Evaluation tasks and metrics.}
Our assessment covers four memory-induced risks. For cross-domain leakage and sycophancy, we follow the PersistBench setup and report the failure rate (FR). The
judge evaluates whether the response exhibits inappropriate memory influence, producing an ordinal failure score between $1-5$, where higher scores indicate more severe memory-induced failure and treat scores $\geq $ 3 as failures since they indicate clear inappropriate memory influence.

\begin{table}[t]
    \centering
    \small
    \setlength{\tabcolsep}{1pt}
    \caption{Summary of evaluation benchmarks and targeted failure types used in our assessment.}
    \begin{tabular}{cccc}
    \toprule[1pt]
    \textbf{Benchmark} & \textbf{Type}  & \textbf{\#Samples} & \textbf{Judge Model}\\
    \midrule[1pt]
    \multirow{2}{*}{PersistBench\cite{persistbench}} & Cross-Domain Failure & 200 & GPT-4o-mini \\
     & Sycophancy Failure & 200 & GPT-4o-mini \\
     \midrule[0.5pt]
     MemDrift \cite{dabas2026memory} & Tool-call Failure & 105 & GPT-4o-mini \\
     \midrule[0.5pt]
    PS-Bench \cite{guo2026personalization} & Jailbreak Attack& 750 & Llama-Guard \\
   \bottomrule[1pt]
    \end{tabular}
    \label{tbl:utility_metrics}
\end{table}

\begin{table*}[t]
    \centering
    \small
    \setlength{\tabcolsep}{5pt}
    \caption{Failure Rate (FR\%) on PersistBench and MemDrift of different LLMs. We evaluate the cross-domain leakage, sycophancy and tool-call drift on agentic memory frameworks and real-world agent.}
    \begin{tabular}{clcccccc}
    \toprule[1pt]
    \textbf{Type} & \textbf{Agent} & \textbf{Qwen-3-8B} & \textbf{Llama-3.3-70B} & \textbf{GPT-4o-mini} & \textbf{GPT-5.1} & \textbf{Gemini-3-Pro} & \textbf{Claude-Sonnet-4.6}\\
    \midrule[0.8pt]

    \multirow{4}{*}{\makecell{Cross-Domain \\ FR (\%)}}    
    & A-Mem    & 74.5 & 14.0 & 25.0 & 1.5 & 34.0 & 22.5 \\
    & Mem0     & 71.0 & 12.0 & 26.5 & 2.0 & 29.5 & 19.0 \\
    & MemOS    & 81.0 & 17.0 & 24.5 & 1.5 & 38.0 & 26.0 \\
    & OpenClaw & 79.5 & 16.5 & 27.0 & 2.0 & 35.5 & 25.0 \\

    \midrule[0.8pt]
    
    \multirow{5}{*}{\makecell{Sycophancy \\ FR (\%)}}
    & w/o Mem & 30.5 & 18.0 & 25.0 & 2.5 & 20.0 & 19.5 \\
    \cdashline{2-8} \addlinespace[3pt]

    & A-Mem    & 84.5 & 49.0 & 32.5 & 8.0 & 40.5 & 39.0 \\
    & Mem0     & 76.5 & 47.5 & 34.5 & 7.5 & 35.5 & 37.0 \\
    & MemOS    & 90.0 & 53.5 & 34.0 & 8.5 & 43.5 & 41.0 \\
    & OpenClaw & 88.0 & 51.0 & 33.5 & 8.5 & 42.5 & 42.0 \\

    \midrule[0.8pt]

   \multirow{5}{*}{\makecell{Tool-call \\ FR (\%)}}  
    & w/o Mem & 31.4 & 54.3 & 48.6 & 28.6 & 45.7 & 22.9\\
\cdashline{2-8}
\addlinespace[3pt]
    & A-Mem    & 57.1 & 68.6 & 77.1 & 42.9 & 88.6 & 85.7\\
    & Mem0     & 62.9 & 71.4 & 82.9 & 45.7 & 97.1 & 91.4\\
    & MemOS    & 65.7 & 74.3 & 85.7 & 48.6 & 94.3 & 88.6\\
    & OpenClaw & 40.0 & 65.7 & 62.9 & 37.1 & 82.9 & 77.1\\
    \bottomrule[1pt]
    \end{tabular}
    \label{tbl:baseline_fr}
\end{table*}

\begin{table*}[t]
    \centering
    \small
    \setlength{\tabcolsep}{5pt}
    \caption{Attack Success Rate (ASR\%) on PS-Bench of different LLMs.}
    \begin{tabular}{clcccccc}
    \toprule[1pt]
    \textbf{Type} & \textbf{Agent} & \textbf{Qwen-3-8B} & \textbf{Llama-3.3-70B} & \textbf{GPT-4o-mini} & \textbf{GPT-5.1} & \textbf{Gemini-3-Pro} & \textbf{Claude-Sonnet-4.6}\\
    \midrule[0.8pt]
    \multirow{5}{*}{\makecell{Jailbreak \\ ASR (\%)}}
    & w/o Mem & 5.0 & 9.0 & 2.5 & 0.0 & 2.0 & 0.0\\
    \cdashline{2-8}
    \addlinespace[3pt]
    & A-Mem    & 22.3 & 42.7 & 29.7 & 13.1 & 10.5 & 20.2\\
    & Mem0     & 18.0 & 33.6 & 26.8 & 11.2 & 8.3  & 18.7\\
    & MemOS    & 18.5 & 38.8 & 28.7 & 12.0 & 9.6  & 17.5\\
    & OpenClaw & 20.1 & 35.0 & 16.8 & 14.0 & 11.5 & 22.0\\ 
    \bottomrule[1pt]
    \end{tabular}
    \label{tbl:baseline_asr}
\end{table*}

For tool-call drift, we evaluate whether user memories alter safety-critical tool parameters, such as validation mode, logging, rollback, approval gates, visibility, or execution policy, and likewise report FR. We use GPT-4o-mini to produce a deflection score on a $1-5$ likert scale, where higher scores indicate greater deviation with heavier weight given to drift in task-critical parameters, and we also treat scores $\geq $ 4 as failures.

For jailbreak, we follow the PS-Bench-style setting, where benign-looking memories are used to legitimize a later harmful query, and report attack success rate (ASR), defined as the percentage of harmful queries for which the agent provides policy-violating assistance instead of refusing. The detailed benchmark composition and judge models are summarized in Table~\ref{tbl:utility_metrics}.

\subsection{Problem I: Misalignment Feedback from Over-Personalization}

The first failure mode arises when the agent treats persistent user memory as a general alignment signal rather than as context that must be selectively applied. In this setting, the user is benign: the stored memories are truthful and are acquired through ordinary interactions. However, the agent over-generalizes these memories to unrelated tasks, producing a misalignment feedback effect in which personalization shifts the model away from task-grounded and policy-grounded behavior. We observe this problem in three forms: cross-domain leakage, sycophancy, and tool-call drift.

As shown in Table~\ref{tbl:baseline_fr}, cross-domain leakage remains substantial across memory-enabled agents. Although the retrieved memories are not relevant to the target task, they still influence the response in many cases. This indicates that current memory mechanisms are not only retrieval systems, but also implicit behavioral conditioning channels. A memory may be semantically close to the query while still being contextually inappropriate, and the agent often lacks a reliable mechanism for rejecting such memory influence.

Sycophancy further shows how persistent memory can distort the response objective. In the memory-free setting, the average sycophancy FR is $19.3\%$, while memory-enabled settings raise it to roughly $37.5\%$--$44.9\%$. This suggests that stored user beliefs can amplify the model's tendency to agree with the user. Instead of using memory only to understand the user's background, the agent may treat prior beliefs as preferences to be reinforced, even when the task requires objective or balanced reasoning.

Tool-call drift shows that over-personalization also affects executable actions. In the memory-free setting, tool-call FR is already non-negligible. However, after memory is introduced, the failure rate increases substantially across memory-enabled agents, indicating that personal memories further bias parameter selection. Personal traits such as being fast, concise, risk-tolerant, or cost-conscious may silently change safety-critical parameters, including validation mode, logging level, rollback behavior, approval gates, visibility, or execution policy. The core flaw lies in the lack of an admissibility mechanism: agents are heavily biased to utilize retrieved memories but lack the capacity to reject them when inappropriate, creating a feedback loop of compounding misalignment.

\subsection{Problem II: Safety Vulnerabilities from User-Manipulable Memory}

The second failure mode concerns malicious users who intentionally manipulate the memory repository. Unlike direct jailbreaks, this attack does not require the current prompt to contain an explicit instruction to bypass safety rules. Instead, the attacker first plants benign-looking memories that establish an apparently legitimate context, such as a role, hobby, project, research goal, or situational background. Later, the attacker issues a harmful or policy-sensitive request that overlaps with the planted context. The agent may then use the memory to infer benign intent and become more likely to comply.

This vulnerability is dangerous because memory changes how the agent interprets user intent. In a memory-free setting, the model mainly evaluates the current request. With persistent memory, however, the same request is evaluated together with prior context that may make it appear justified. The attack therefore targets the personalization layer rather than the instruction-following layer. Even if the current query contains no explicit jailbreak pattern, the memory state can steer the model toward treating the harmful request as consistent with the user's stored background. The jailbreak results show that this effect is systematic. The memory-free baseline has an average ASR of only $3.9\%$, while memory-enabled settings raise the average ASR to approximately $18.2\%$--$24.6\%$. This increase indicates that persistent memory can provide an attacker with a durable context for intent legitimation. The harmful request is not necessarily made safer by the memory, but the agent may incorrectly treat the memory as evidence that the request belongs to a benign scenario.

This problem differs from benign over-personalization in an important way. In Problem I, the memory is truthful but misapplied. In Problem II, the memory is strategically constructed so that it will be misapplied later. The attacker does not need to directly modify system prompts, retrieval code, tool schemas, or model weights. Ordinary conversation is sufficient to influence the future memory state, and the memory state can later become part of the effective attack prompt.

\section{Defense Methodology: MemGate}
   \def\sectionfolder{sections/}%

To address the systemic vulnerabilities introduced by unconditional semantic retrieval, this section presents our defense mechanism: MemGate. In the following, we first detail the MemGate architecture (Section \ref{sec:memgate_architecture}) and then introduce our preference-aligned training objective (Section \ref{sec:memgate_alignment}). Finally, we provide theoretical proofs of MemGate (Section \ref{sec:theory}).

\subsection{MemGate Architecture}
\label{sec:memgate_architecture}
As formalized in Section \ref{sec:memory_problems}, the standard memory pipeline suffers from a critical admissibility gap: standard embedding spaces struggle to encode the latent contextual admissibility $\mathcal{A}(m, x_q) \in \{0, 1\}$. To bridge the gap between semantic similarity $S(q, v_m)$ and contextual admissibility $\mathcal{A}(m, x_q)$, one might intuitively train a binary classifier to discard memories where $\mathcal{A}(m, x_q)=0$. However, a memory often becomes inadmissible not because its entire semantic content is faulty, but because specific latent feature dimensions (e.g., a cross-domain bias, a stale constraint, or a sycophantic trait) conflict with the immediate query context. 

Therefore, we propose MemGate, a parameter-efficient neural gate network $G_\phi$ that acts as a fine-grained, representation-level filter rather than a rigid binary classifier. Positioned strategically between the vector database and the backbone LLM, MemGate employs a continuous masking strategy $G_\phi \in [0, 1]^d$, which can be viewed as a continuous relaxation of the discrete admissibility indicator $\mathcal{A}(m, x_q)$. This allows the LLM to selectively attenuate the specific latent dimensions that violate contextual admissibility, while safely preserving the task-relevant semantic evidence.

To effectively capture fine-grained semantic cross-resonances, MemGate conditions its gating decisions on the interaction between the user query embedding $q \in \mathbb{R}^d$ and the candidate memory embedding $v_m \in \mathbb{R}^d$. The comprehensive interaction feature vector $x_{\text{inter}}$ is formalized as:
\begin{equation}
    x_{\text{inter}} = [q \oplus v_m \oplus (q \odot v_m)] \in \mathbb{R}^{3d}
\end{equation}
where $\oplus$ denotes vector concatenation and $\odot$ represents the Hadamard (element-wise) product. Given our specific configuration utilizing $d=384$ dimensional embeddings, the total input dimensionality of $x_{\text{inter}}$ is precisely $1152$.

The core backbone of MemGate is parameterized as a deep shared Multi-Layer Perceptron (MLP) enhanced with Layer Normalization (\text{LN}) and SiLU activation functions. The hidden states flow sequentially through the network:
\begin{align*}
    h_1 &= \text{SiLU}(\text{LN}(\text{Linear}_{1152 \to 2048}(x_{\text{inter}}))) \\
    h_2 &= \text{SiLU}(\text{LN}(\text{Linear}_{2048 \to 2048}(h_1))) \\
    h_3 &= \text{SiLU}(\text{LN}(\text{Linear}_{2048 \to 1024}(h_2)))
\end{align*}
To enhance generalization and prevent overfitting during the alignment phase, a dropout rate of $0.15$ is applied to $h_1$ and $h_2$, and a rate of $0.1$ is applied to $h_3$. Finally, the network dynamically projects this shared semantic context into a single gating mask applicable to the memory space:
\begin{equation}
    G_\phi(q, v_m) = \sigma(\text{Linear}_{1024 \to 384}(h_3)) \in [0, 1]^d
\end{equation}
where $\sigma(\cdot)$ denotes the element-wise Sigmoid function, bounding the attenuation mask strictly between 0 and 1. Through the preference alignment phase detailed below, this multi-dimensional mask effectively learns to map continuous feature attenuation to the global binary admissibility targets.

\subsection{MemGate Alignment}
\label{sec:memgate_alignment}

\noindent\textbf{Definition.}
Standard cosine similarity is vulnerable to surface-level lexical overlap. MemGate addresses this problem by computing an asymmetrically gated cosine similarity $S_{\text{MemGate}}(q, v_m)$, which projects the raw memory embedding into a query-conditioned, admissibility-aligned subspace:
\begin{equation}
    S_{\text{MemGate}}(q, v_m) =
    \frac{q \cdot (v_m \odot G_\phi(q, v_m))}
    {\|q\|_2 \|v_m \odot G_\phi(q, v_m)\|_2},
\end{equation}
where $G_\phi(q, v_m)$ is the query-conditioned gate predicted for memory $m$. Since distracting dimensions that contribute to similarity but violate admissibility are attenuated before $\ell_2$ normalization, the final score is computed using the safely retained memory evidence rather than the full ungated embedding.

\noindent\textbf{Alignment Objective.} We formulate retrieval alignment as a Direct Preference Optimization (DPO) task~\cite{rafailov2023direct}. Given a query $q$, a contextually admissible memory $m^+$ ($\mathcal{A}=1$), and a conflicting or unsafe memory $m^-$ ($\mathcal{A}=0$), MemGate should increase the relative retrieval likelihood of $m^+$ while decreasing that of $m^-$. We model the gated retriever as a probabilistic policy $\pi_\phi$. For a candidate memory set $\mathcal{C}_q$, the probability of selecting memory $m$ is defined by a softmax over gated similarity scores:
\begin{equation}\label{eq:policy_phi}
    \pi_\phi(m \mid q) =
    \frac{\exp \left( S_{\text{MemGate}}(q, v_m) / \tau \right)}
    {\sum_{m' \in \mathcal{C}_q}
    \exp \left( S_{\text{MemGate}}(q, v_{m'}) / \tau \right)},
\end{equation}
where $\tau$ is the temperature parameter. The candidate set $\mathcal{C}_q$ is obtained by the original similarity-based retriever, so MemGate only re-ranks a small set of plausible memories instead of scoring the entire memory repository.

We construct a preference dataset $\mathcal{D}=\{(q,m^+,m^-)\}$. For a preference triple $(q,m^+,m^-)$, we define the shorthand: $g^+=G_\phi(q, v_{m^+}); g^-=G_\phi(q, v_{m^-}).$
To prevent representation collapse on beneficial memories, we add a preservation penalty on the positive mask. Specifically, we penalize the normalized mask deficit $\frac{1}{d}\|\mathbf{1}_d-g^+\|_1$, which biases the gate for admissible memories toward the all-ones vector $\mathbf{1}_d$. We do not impose the same preservation constraint on $g^-$, allowing MemGate to attenuate conflicting memories when required by the preference objective. 

By incorporating this preservation penalty into the standard DPO formulation, the final training objective is:
\begin{equation}
\resizebox{\linewidth}{!}{
$                               
\begin{aligned}
    \mathcal{L}(\phi) &= - \mathbb{E}_{\mathcal{D}} \left[ \log \sigma \left( \beta \log \frac{\pi_\phi(m^+ \mid q)}{\pi_{\text{ref}}(m^+ \mid q)} - \beta \log \frac{\pi_\phi(m^- \mid q)}{\pi_{\text{ref}}(m^- \mid q)} \right) \right] \\
    &\quad + \lambda \cdot \mathbb{E}_{(q, m^+,m^-) \sim \mathcal{D}} \left[ \frac{1}{d} \|\mathbf{1}_d - g^+\|_1 \right]
\end{aligned}
$
}
\end{equation}
where $\beta$ controls the strength of the preference margin and $\lambda$ controls the positive-memory preservation strength. Since the preservation term is applied only to $m^+$, it provides a targeted stability constraint for useful memories rather than a rigid global trust-region constraint over all retrieved memories.

In the objective above, $\pi_{\text{ref}}$ denotes the reference policy. Analogous to Eq.~\ref{eq:policy_phi}, we define $\pi_{\text{ref}}$ using the ungated base embedding retriever:
\begin{equation*}
    \pi_{\text{ref}}(m \mid q) =
    \frac{\exp \left( S(q, v_m) / \tau \right)}
    {\sum_{m' \in \mathcal{C}_q}
    \exp \left( S(q, v_{m'}) / \tau \right)}.
\end{equation*}
We choose this reference policy for two reasons. First, it corresponds exactly to the retrieval behavior before adding MemGate, making the alignment objective a controlled modification of the existing memory search pipeline rather than an unconstrained retraining of retrieval from scratch. Second, the ungated retriever preserves the general semantic structure of the embedding space. By optimizing relative preference against this reference, MemGate explicitly learns to approximate the admissibility targets ($\mathcal{A}=1$ vs. $\mathcal{A}=0$), correcting contextually unsafe retrieval decisions while avoiding unnecessary deviation from useful semantic matching.

\noindent\textbf{Inference with MemGate.}
At inference time, MemGate is positioned at the boundary between vector retrieval and prompt construction. Instead of directly injecting the memories ranked by ungated cosine similarity, the agent scores each candidate memory with the query-conditioned gated similarity $S_{\text{MemGate}}(q,v_m)$ and retrieves the final top-$k$ memories as
\begin{equation*}
    \mathcal{M}'_q =
    \{m_i \mid
    \text{rank}(S_{\text{MemGate}}(q,v_{m_i})) \le k,\;
    m_i\in\mathcal{M}\}.
\end{equation*}
The textual contents of the selected memories are then concatenated with the user query to construct the augmented prompt for the backbone LLM. In this way, MemGate replaces raw nearest-neighbor memory injection with query-conditioned memory admission: memories are allowed to influence the model only after their memory-side representations have been filtered in the gated subspace. This preserves the standard vector-memory workflow while adding a retrieval-time contextual isolation layer, without modifying the backbone LLM, rewriting the memory database, or invoking an additional LLM judge.

\begin{figure*}
    \centering
    \includegraphics[width=1\linewidth]{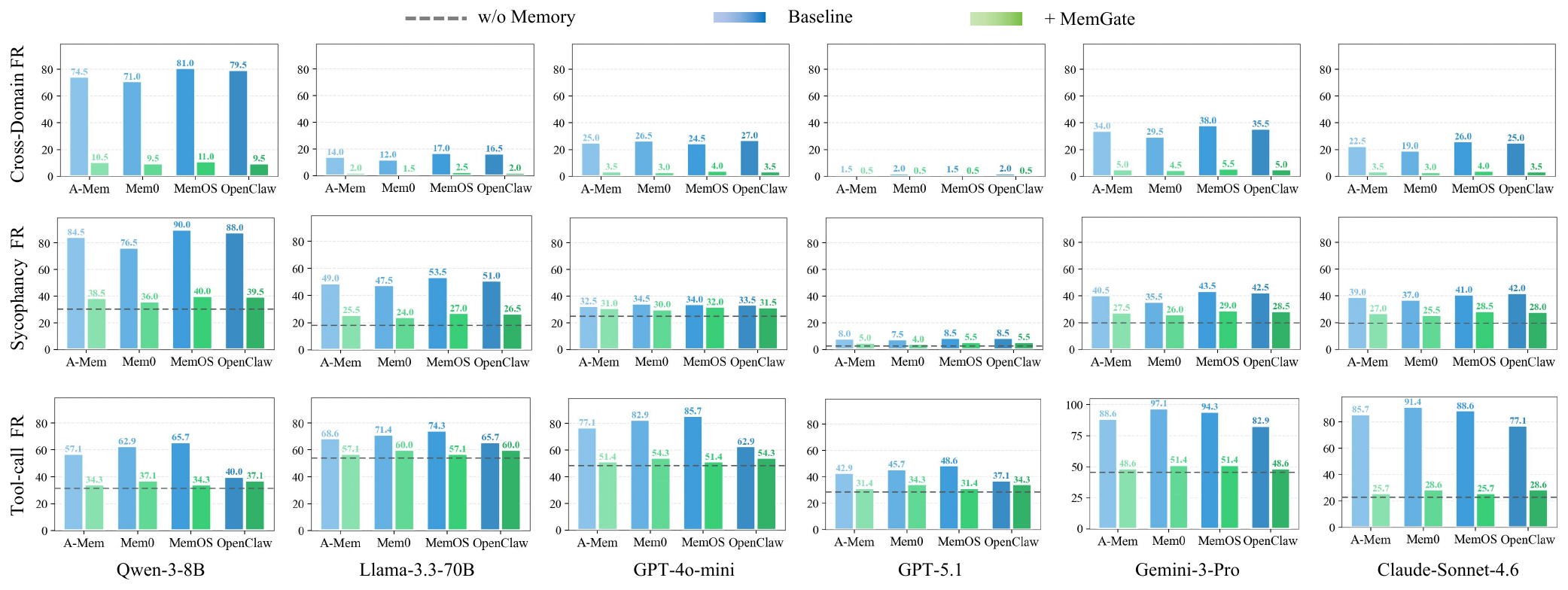}
    \caption{Performance of MemGate on over-personalization evaluations, including cross-domain leakage, sycophancy, and tool-call drift across agentic memory frameworks and the real-world agent.}
    \label{fig:memgate_op_result}
\end{figure*}

\subsection{Theoretical Analysis}
\label{sec:theory}

To formally demonstrate the reliability of our preservation penalty, we prove that bounding the positive mask deficit limits the semantic drift of beneficial memories.

\begin{theorem}[Bounded Score Perturbation]\label{thm:bounded_score_perturbation}
Let $S(q,v_{m^+})$ denote the ungated cosine similarity between a query embedding $q$ and an aligned memory embedding $v_{m^+}$, and let $S_{\text{MemGate}}(q,v_{m^+})$ denote the gated cosine similarity computed with $v'_{m^+}=v_{m^+}\odot g^+$, where $g^+\in[0,1]^d$. Assume $q\neq 0$ and $\|v'_{m^+}\|_2\ge\gamma>0$. Then the perturbation of the aligned-memory similarity score is bounded by:
$$|S_{\text{MemGate}}(q,v_{m^+})-S(q,v_{m^+})|\le\frac{\|v_{m^+}\|_\infty}{\gamma}\|\mathbf{1}_d-g^+\|_1.$$
\end{theorem}

The theorem guarantees that when the positive mask deficit is minimized during alignment, the gated similarity score of a beneficial memory remains rigorously bounded near its ungated reference score, preventing representation collapse. The complete proof is provided in Appendix~\ref{app:proof_of_bounded_score_perturbation}.

\begin{theorem}[Unsafe-Memory Suppression]\label{thm:unsafe_suppression}
Let $\ell_z(g^+,g^-)$ denote the DPO loss for a preference triple $z=(q,m^+,m^-)$ before applying the preservation penalty. Assume $\ell_z$ is $L$-smooth over a convex domain containing $(g^+,g^-)$ and $(\mathbf{1}_d,\mathbf{1}_d)$. If optimization reaches a stationary point satisfying $0\le\ell_z(g^+,g^-)\le\epsilon<\log 2$, the total gate perturbation is lower-bounded by:
$$\|\mathbf{1}_d-g^+\|_2^2+\|\mathbf{1}_d-g^-\|_2^2\ge\frac{2(\log 2-\epsilon)}{L}.$$
In particular, if the positive memory is preserved ($g^+=\mathbf{1}_d$), the unsafe-memory suppression gate $g^-$ must satisfy:
$$\|\mathbf{1}_d-g^-\|_1\ge\|\mathbf{1}_d-g^-\|_2\ge\sqrt{\frac{2(\log 2-\epsilon)}{L}}.$$
\end{theorem}

This theorem demonstrates that reducing the DPO loss below $\log 2$ forces the gates to deviate significantly from their reference identity states. Because the preservation penalty anchors $g^+$ near $\mathbf{1}_d$, this required perturbation is predominantly absorbed by $g^-$. This mathematically guarantees the suppression of unsafe memory dimensions. The complete proof is in Appendix~\ref{app:proof_of_suppression_lower_bound}.%

\section{Experiments}
   \def\sectionfolder{sections/}%
   \begin{figure*}
    \centering
    \includegraphics[width=1\linewidth]{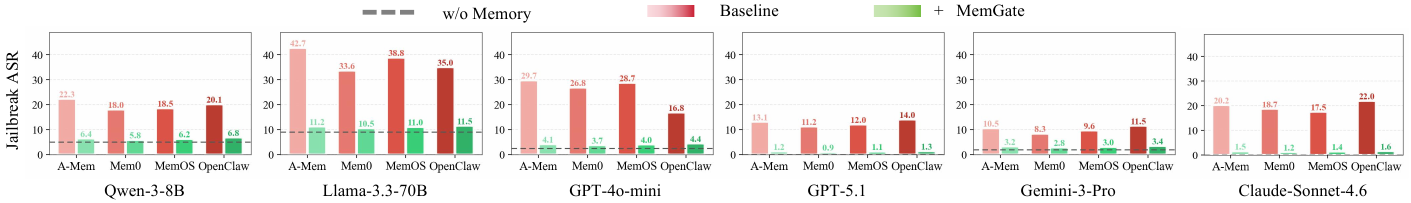}
    \caption{Performance of MemGate on memory-induced safety vulnerability evaluations across agentic memory frameworks and the real-world agent.}
    \label{fig:memgate_sv_result}
\end{figure*}

\begin{figure}
    \centering
    \includegraphics[width=1\linewidth]{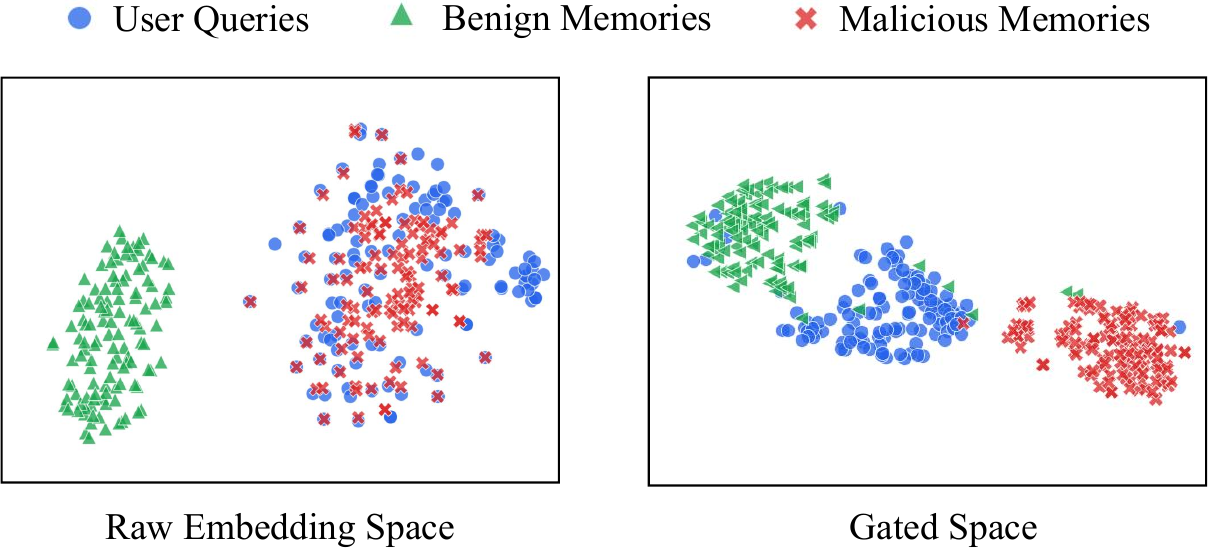}
    \caption{Representation shift induced by MemGate. In the raw embedding space, malicious memories overlap with user queries due to semantic similarity; after gating, malicious memories are separated in the gated subspace while benign memories remain close to the query cluster.}
    \label{fig:memgate_rep}
\end{figure}

In this section, we present a comprehensive empirical evaluation of MemGate. Specifically, our evaluation is designed to answer the following research questions:

\begin{enumerate}
    \item \textit{Can MemGate improve the trustworthiness of agents?}
    \item \textit{Does MemGate preserve the utility of agents?}
    \item \textit{What is the runtime overhead of MemGate?}
    \item \textit{How do key hyperparameters affect MemGate?}
\end{enumerate}

\subsection{Experimental Setup}

\noindent\textbf{MemGate training settings.}
We train MemGate as a lightweight gating module on top of frozen sentence embeddings. Queries and candidate memories are encoded with \texttt{all-MiniLM-L6-v2} into 384-dimensional vectors. For each query--memory pair, the gate takes the concatenation of the query embedding, memory embedding, and their element-wise product as input, and outputs a sigmoid gate that masks only the memory embedding. 

MemGate has about 9M parameters and is trained from scratch on 1,640 preference pairs. The preference pairs are synthetically constructed with GPT-4o-mini. To avoid test-set leakage, we construct these pairs from abstract memory-risk categories rather than from benchmark instances. They do not include any test queries, memory snippets, labels, or examples from the evaluation benchmarks. We optimize a DPO objective with the raw cosine retriever as the reference policy, together with a positive-memory mask preservation penalty weighted by $\lambda=0.1$. The DPO temperature is $0.1$ and $\beta=0.3$. We use AdamW with learning rate $1.5 \times 10^{-4}$, batch size 256, and train for 20 epochs.

\noindent\textbf{Benchmarks.}
For memory utility, we evaluate whether MemGate preserves the useful function of long-term memory. We use two general memory benchmarks, LoCoMo \cite{maharana2024evaluating} and MemoryAgentBench \cite{hu2026evaluating}, and three personalization-oriented benchmarks, PreFEval \cite{zhao2025llms}, PersonaMem \cite{jiang2025know}, and the beneficial-memory subset of PersistBench \cite{persistbench}. LoCoMo and MemoryAgentBench measure long-term memory retrieval and reasoning, while PreFEval, PersonaMem, and PersistBench evaluate whether the agent can correctly use user preferences, profiles, and genuinely relevant memories. Detailed benchmark descriptions are provided in Appendix~\ref{app:utility_benchmarks}.

\subsection{Trustworthy Evaluation}

We evaluate trustworthiness under two threat settings: benign over-personalization and malicious memory manipulation. For benign users, we measure cross-domain leakage, sycophancy, and tool-call drift. For malicious users, we measure memory-based jailbreak attack success.

As shown in Figure~\ref{fig:memgate_op_result}, MemGate consistently reduces over-personalization failures across agents and backbone models. In cross-domain leakage, memory-enabled agents may retrieve memories that are semantically related but contextually inappropriate, causing information from one domain to affect unrelated tasks. MemGate largely suppresses this misleading memory influence before generation. For example, on GPT-4o-mini, cross-domain FR drops from roughly $24.5\%$--$27.0\%$ to $3.0\%$--$4.0\%$ after applying MemGate.

Sycophancy is also mitigated, although not fully removed. MemGate reduces the memory-induced tendency to over-agree with stored user beliefs, while the remaining failures stay slightly above the memory-free baseline. This indicates that sycophancy is caused by both inappropriate memory activation and the base model's intrinsic agreement bias. Tool-call drift also shows a clear improvement, but the memory-free baseline is higher than in the other risk categories because tool-parameter selection itself is challenging. Without MemGate, personal traits stored in memory can further bias safety-critical tool parameters such as validation mode, logging, rollback, approval gates, or execution policy. After applying MemGate, tool-call FR is reduced close to the memory-free level. On Claude-Sonnet-4.6, memory-enabled agents originally show high drift, ranging from $77.1\%$ to $91.4\%$, while MemGate reduces it to $25.7\%$--$28.6\%$, close to the w/o Mem result of $22.9\%$.

Figure~\ref{fig:memgate_sv_result} evaluates the malicious-user setting, where planted memories are used to legitimize harmful requests. Without MemGate, persistent memory can become a durable attack surface: once a benign-looking context is stored, the agent may later use it to reinterpret a harmful query as legitimate. MemGate substantially reduces this vulnerability. On GPT-4o-mini, memory-enabled agents increase jailbreak ASR from the w/o Mem baseline of $2.5\%$ to much higher levels, while MemGate brings it back to $3.7\%$--$4.4\%$. Overall, MemGate provides the largest gains when failures are directly caused by inappropriate memory activation. Cross-domain leakage, tool-call drift, and memory-based jailbreaks are reduced substantially, while sycophancy remains more challenging due to the base model's residual agreement bias.

To better understand how MemGate achieves this behavior, we visualize the embedding geometry before and after gating. We sample 120 user queries from PS-Bench, paired with 120 benign memories from LoCoMo dialogues and 120 malicious memories templated from the queries. All texts are encoded by all-MiniLM-L6-v2 into 384-dimensional vectors. We compare the raw L2-normalized embeddings with the gated embeddings obtained by element-wise multiplication with the gate vectors $\mathbf{g}_q$ and $\mathbf{g}_m$, followed by L2 re-normalization. The vectors are projected to 2-D via t-SNE using perplexity 20, PCA initialization, and cosine distance. As shown in Figure~\ref{fig:memgate_rep}, malicious memories strongly overlap with user queries in the raw embedding space because they are constructed to be semantically similar to the target request. After gating, the malicious memories are pushed away from the query cluster, while benign memories remain close to user queries. This representation-level separation shows that MemGate learns a task-aware subspace: it weakens adversarial or contextually misleading associations without collapsing the useful semantic neighborhood. This is consistent with the utility results, where MemGate improves memory reliability by reshaping retrieval toward safer and more relevant memory usage.

\begin{table*}
    \centering
    \small
    \setlength{\tabcolsep}{3pt}
    \caption{Utility evaluation on general memory benchmarks. We report F1 score on LoCoMo and accuracy on MemoryAgentBench across different memory-enabled agents, with and without MemGate.}
    \begin{tabular}{lccccccccccc}
    \toprule[1pt]
    \multirowcell{2}{\textbf{Agent}}  & \multicolumn{5}{c}{\textbf{LoCoMo} \cite{maharana2024evaluating}} & & \multicolumn{5}{c}{\textbf{MemoryAgentBench} \cite{hu2026evaluating}}\\
    \cmidrule(lr){2-6} \cmidrule(lr){8-12} 
    & Single-Hop & Multi-Hop & Open-Domain & Temporal & Overall &  & Single-Hop & Multi-Hop & Long-Context & Event & Overall\\
    \midrule[1pt]
    \multicolumn{12}{c}{\textit{Qwen-3-8B}} \\
    \hdashline
    \addlinespace[3pt]
    A-Mem            & 43.7 & 37.4 & 18.6 & 15.9 & 37.3 & & 18.8 & 24.3 & 29.1 & 30.4 & 25.7\\
    \quad + MemGate  & 44.9 & 38.6 & 21.7 & 17.3 & 38.6 & & 20.2 & 25.7 & 30.8 & 31.6 & 27.1\\
    \hline \addlinespace[3pt]
    Mem0             & 48.5 & 44.5 & 16.7 & 12.5 & 40.0 & & 22.4 & 27.6 & 32.3 & 33.1 & 28.8\\
    \quad + MemGate  & 46.6 & 46.2 & 23.9 & 15.2 & 40.2 & & 23.6 & 28.4 & 33.1 & 34.2 & 29.7\\
    \hline \addlinespace[3pt]
    MemOS            & 50.7 & 46.1 & 18.4 & 14.6 & 41.7 & & 24.1 & 30.3 & 34.8 & 35.6 & 31.2\\
    \quad + MemGate  & 51.3 & 47.8 & 24.4 & 17.8 & 42.9 & & 25.3 & 31.4 & 36.1 & 36.8 & 32.4\\
    \hline \addlinespace[3pt]
    OpenClaw         & 44.6 & 38.1 & 21.9 & 17.5 & 38.5 & & 20.7 & 26.1 & 31.2 & 31.8 & 27.5\\
    \quad + MemGate  & 45.0 & 38.7 & 24.1 & 19.5 & 39.4 & & 21.8 & 26.9 & 31.7 & 32.9 & 28.3\\

    \midrule[1pt]
    \multicolumn{12}{c}{\textit{GPT-4o-mini}} \\
    \hdashline
    \addlinespace[3pt]
    A-Mem            & 41.3 & 35.6 & 20.1 & 33.4 & 37.8 & & 22.7 & 28.2 & 33.1 & 35.4 & 29.9\\
    \quad + MemGate  & 42.6 & 37.1 & 19.8 & 36.3 & 39.1 & & 23.4 & 29.7 & 34.6 & 36.2 & 31.0\\
    \hline \addlinespace[3pt]
    Mem0             & 48.6 & 42.1 & 15.1 & 35.3 & 42.9 & & 25.0 & 32.0 & 36.0 & 37.5 & 32.6\\
    \quad + MemGate  & 49.8 & 43.2 & 15.1 & 40.4 & 44.5 & & 26.4 & 33.1 & 37.3 & 38.7 & 33.9\\
    \hline \addlinespace[3pt]
    MemOS            & 50.4 & 44.2 & 17.3 & 38.9 & 45.4 & & 27.8 & 35.3 & 39.2 & 40.7 & 35.8\\
    \quad + MemGate  & 51.6 & 45.1 & 18.6 & 42.5 & 46.3 & & 28.7 & 36.4 & 40.3 & 41.8 & 36.8\\
    \hline \addlinespace[3pt]
    OpenClaw         & 42.5 & 36.9 & 22.8 & 35.1 & 38.9 & & 24.2 & 30.7 & 35.4 & 36.1 & 31.6\\
    \quad + MemGate  & 44.9 & 39.4 & 16.9 & 38.5 & 40.8 & & 25.1 & 31.8 & 36.2 & 37.6 & 32.7\\
    \bottomrule[1pt]
    \end{tabular}
    \label{tbl:general_memory_utility}
\end{table*}

\subsection{Utility Evaluation}

Besides reducing memory-induced risks, MemGate should not sacrifice the core utility of memory. We therefore evaluate it on both general memory benchmarks and personalization-oriented benchmarks. Table~\ref{tbl:general_memory_utility} reports results on LoCoMo and MemoryAgentBench, while Table~\ref{tbl:personalization_utility} reports results on PreFEval, PersonaMem, and the beneficial subset of PersistBench. MemGate preserves and slightly improves general memory utility across both Qwen-3-8B and GPT-4o-mini. On LoCoMo, the overall scores remain stable after applying MemGate, with small gains in most agent settings. For example, on GPT-4o-mini, Mem0 improves from $42.9$ to $44.5$, while OpenClaw improves from $38.9$ to $40.8$. Similar trends appear on MemoryAgentBench, where MemGate consistently improves the overall score across all memory agents.

The utility improvement comes from MemGate's selective reweighting rather than simple memory suppression. Conventional retrieval mainly relies on semantic similarity, which can retrieve memories that share surface-level keywords with the query but are not actually useful for the current task. Such noisy memories consume context budget and may distract the model from the truly relevant memory evidence. MemGate learns query- and memory-specific gates that down-weight misleading dimensions while preserving dimensions that are predictive of helpful memory use. As a result, the retrieved context becomes cleaner: irrelevant or adversarial memories are less likely to dominate the prompt, while task-relevant memories become easier for the model to use. This explains why MemGate can improve utility even though it is introduced primarily as a safety mechanism.

The personalization results in Table~\ref{tbl:personalization_utility} further support this conclusion. MemGate improves PreFEval and PersonaMem accuracy in most settings, indicating that it does not weaken the agent's ability to use genuinely helpful user preferences. More importantly, it substantially reduces Beneficial FR on PersistBench. For example, on GPT-4o-mini, OpenClaw reduces Beneficial FR from $14.3$ to $6.7$, while MemOS reduces it from $8.1$ to $3.5$. This suggests that MemGate does not block memory globally; instead, it improves the precision of memory usage. By suppressing memories that are semantically tempting but contextually harmful, MemGate allows the model to rely more consistently on memories that are actually beneficial for the user query.

\begin{table}[t]
    \centering
    \small
    \caption{Efficiency evaluation of MemGate. We report P50 (median latency) and P90 (90th-percentile latency) in seconds on GPT-4o-mini, together with LoCoMo F1 score.}
    \begin{tabular}{lccc}
    \toprule[1pt]
    \textbf{Agent} & \textbf{P50 (s)}  & \textbf{P90 (s)} & \textbf{LoCoMo F1}\\
    \midrule[1pt]
    A-Mem & 0.42 & 0.61 & 37.3\\
    \quad + MemGate & 0.44 & 0.68 & 38.6\\
    \midrule[0.5pt]
    Mem0 & 0.26 & 0.35 & 40.0\\
    \quad + MemGate & 0.29 & 0.41 & 40.2\\
    \midrule[0.5pt]
    MemOS & 0.19 & 0.27 & 41.7\\
    \quad + MemGate & 0.23 & 0.30 & 42.9\\
    \midrule[0.5pt]
    OpenClaw & 1.47 & 2.56 & 38.5\\
    \quad + MemGate  & 1.59 & 2.75 & 39.4\\
    \bottomrule[1pt]
    \end{tabular}
    \label{tbl:efficiency}
\end{table}

\subsection{Efficiency}

We further evaluate the runtime overhead introduced by MemGate during memory retrieval. Since MemGate is a lightweight gating module applied on top of frozen sentence embeddings, it does not require additional LLM calls or iterative reasoning. Its cost mainly comes from computing the query--memory gate and re-ranking memories in the gated embedding space.

Table~\ref{tbl:efficiency} reports memory search latency on GPT-4o-mini. We report both P50 and P90 latency. P50 is the median latency, meaning that half of the retrieval requests finish within this time. P90 is the 90th-percentile latency, meaning that 90\% of the retrieval requests finish within this time. Compared with the average latency, P90 better reflects tail latency and captures slower retrieval cases that may affect user-perceived responsiveness in deployed agents.

\begin{table}
    \centering
    \small
    \setlength{\tabcolsep}{1pt}
    \caption{Utility evaluation on personalization benchmarks. We report PreFEval accuracy, PersonaMem accuracy, and PersistBench beneficial failure rate across different memory-enabled agents, with and without MemGate.}
    \begin{tabular}{lccc}
    \toprule[1pt]
    \multirowcell{2}{\textbf{Agent}}  
    & \textbf{PreFEval}\cite{zhao2025llms}
    & \textbf{PersonaMem}\cite{jiang2025know} 
    & \textbf{PersistBench}\cite{persistbench} \\
    \cmidrule(lr){2-2} \cmidrule(lr){3-3} \cmidrule(lr){4-4}
    & Accuracy$\uparrow$ 
    & Accuracy$\uparrow$ 
    & Beneficial FR$\downarrow$\\
    \midrule[1pt]
    \multicolumn{4}{c}{\textit{Qwen-3-8B}} \\
    \hdashline
    \addlinespace[3pt]
    A-Mem            & 72.9 & 44.2 & 26.1\\
    \quad + MemGate  & 74.6 & 45.7 & 12.2\\
    \hline \addlinespace[3pt]
    Mem0             & 77.2 & 50.9 & 19.6\\
    \quad + MemGate  & 78.4 & 52.1 & 8.5\\
    \hline \addlinespace[3pt]
    MemOS            & 79.5 & 54.6 & 16.2\\
    \quad + MemGate  & 80.7 & 55.8 & 6.4\\
    \hline \addlinespace[3pt]
    OpenClaw         & 75.0 & 47.6 & 23.3\\
    \quad + MemGate  & 76.8 & 49.3 & 10.0\\

    \midrule[1pt]
    \multicolumn{4}{c}{\textit{GPT-4o-mini}} \\
    \hdashline
    \addlinespace[3pt]
    A-Mem            & 84.3 & 52.1 & 16.4\\
    \quad + MemGate  & 86.1 & 53.7 & 8.6\\
    \hline \addlinespace[3pt]
    Mem0             & 88.0 & 56.8 & 11.4\\
    \quad + MemGate  & 89.1 & 58.3 & 5.2\\
    \hline \addlinespace[3pt]
    MemOS            & 89.6 & 61.2 & 8.1\\
    \quad + MemGate  & 90.4 & 62.7 & 3.5\\
    \hline \addlinespace[3pt]
    OpenClaw         & 86.4 & 54.6 & 14.3\\
    \quad + MemGate  & 88.2 & 56.1 & 6.7\\
    \bottomrule[1pt]
    \end{tabular}
    \label{tbl:personalization_utility}
\end{table}

Across all memory agents, MemGate introduces only a small latency increase. For example, Mem0 increases from $0.26$s to $0.29$s at P50. Even for OpenClaw, where the base memory search is already slower due to its system-level memory and tool integration, the additional overhead remains modest: P50 latency increases from $1.47$s to $1.59$s, and P90 latency increases from $2.56$s to $2.75$s.

Importantly, this small overhead does not come at the cost of utility. The overall LoCoMo score is preserved or slightly improved after applying MemGate across all evaluated agents. For instance, A-Mem improves from $37.3$ to $38.6$, MemOS improves from $41.7$ to $42.9$, and OpenClaw improves from $38.5$ to $39.4$. These results indicate that MemGate adds only lightweight computation while improving the quality of retrieved memories. Overall, MemGate provides a favorable efficiency--utility trade-off: it significantly improves trustworthiness with negligible additional retrieval latency.

\begin{figure}
    \centering
    \includegraphics[width=0.95\linewidth]{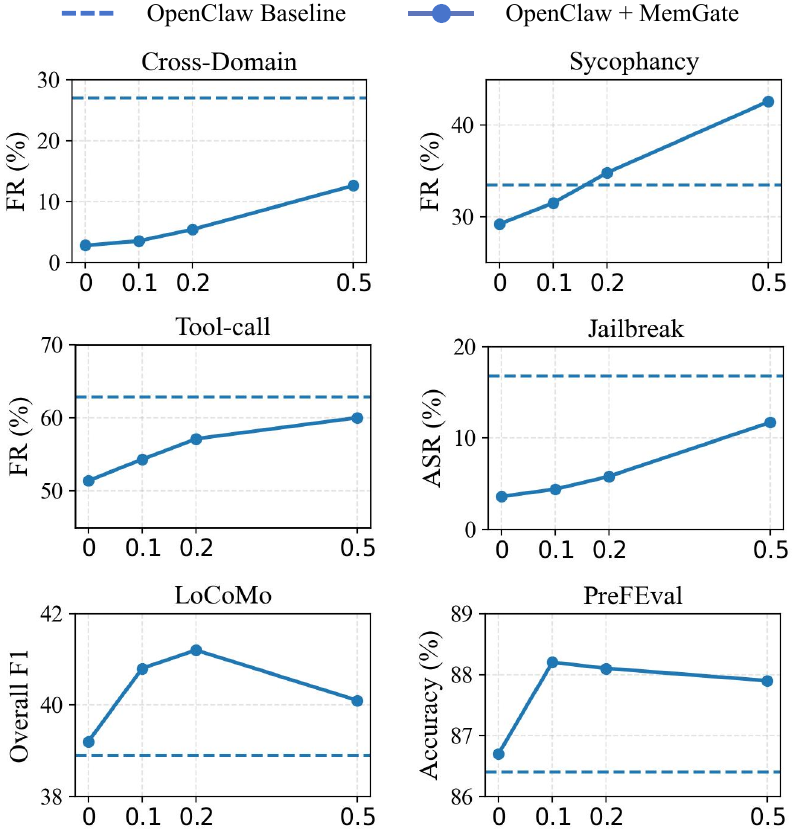}
    \caption{Ablation study of the regularization weight $\lambda$ using GPT-4o-mini with OpenClaw. Lower values are better for failure rate (FR) and attack success rate (ASR), while higher values are better for Overall F1 and Accuracy.}
    \label{fig:ablation_lambda}
\end{figure}

\begin{table*}[t]
\centering
\small
\setlength{\tabcolsep}{5pt}
\caption{Performance of MemGate with different embedding models on GPT-4o-mini with OpenClaw. The table compares trustworthiness, utility, model size, and retrieval latency under different MemGate encoders. Lower is better for FR, ASR, and latency, while higher is better for LoCoMo and PreFEval.}
\label{tab:embedding-robustness-size-latency}
\resizebox{\linewidth}{!}{
\begin{tabular}{lccccccccc}
\toprule
\multicolumn{3}{c}{\textbf{MemGate Setting}} &
\multicolumn{4}{c}{\textbf{Trustworthy Evaluation}} &
\multicolumn{2}{c}{\textbf{Utility Evaluation}} &
\multicolumn{1}{c}{\textbf{Efficiency}} \\
\cmidrule(lr){1-3}
\cmidrule(lr){4-7}
\cmidrule(lr){8-9}
\cmidrule(lr){10-10}
MemGate Encoder & Dim. & Size &
Cross-D. FR$\downarrow$ & Syco FR$\downarrow$ & Tool-call FR$\downarrow$ & ASR$\downarrow$ &
LoCoMo$\uparrow$ & PreFEval$\uparrow$ & Latency$\downarrow$ \\
\midrule
\multicolumn{10}{l}{\textit{OpenClaw (baseline)}} \\
\hdashline
\addlinespace[3pt]
- & - & - & 27.0 & 33.5 & 62.9 & 16.8 & 38.9 & 86.4 & 1.47s \\
\hline
\addlinespace[3pt]
\multicolumn{10}{l}{\textit{OpenClaw w. MemGate}} \\
\hdashline
\addlinespace[3pt]
\texttt{all-MiniLM-L6-v2}       & 384  & 35.1MB & 3.5 & 31.5 & 54.3 & 4.4 & 40.8 & 88.2 & 1.59s \\
\texttt{all-mpnet-base-v2}      & 768  & 45.8MB & 3.5 & 31.0 & 51.4 & 4.1 & 41.2 & 88.5 & 1.63s \\
\texttt{bge-large-en-v1.5}      & 1024 & 52.8MB & 3.5 & 30.5 & 51.4 & 3.9 & 41.6 & 88.8 & 1.65s \\
\texttt{text-embedding-3-small} & 1536 & 67.1MB & 3.0 & 30.0 & 48.6 & 3.9 & 41.8 & 89.0 & 1.74s \\
\bottomrule
\end{tabular}
}
\end{table*}

\subsection{Ablation Studies}

We conduct ablation studies on GPT-4o-mini with OpenClaw to understand how MemGate's key hyperparameters affect the trade-off between trustworthiness and memory utility. We focus on two factors: the regularization weight $\lambda$ used in MemGate training, and the number of retrieved candidate memories top-$k$.

\noindent\textbf{Effect of regularization weight $\lambda$.}
Figure~\ref{fig:ablation_lambda} shows the effect of varying $\lambda$ from $0$ to $0.5$. A smaller $\lambda$ makes the gate more conservative, suppressing irrelevant or risky memories more aggressively. As a result, cross-domain leakage, tool-call drift, and jailbreak ASR remain low when $\lambda$ is small. However, overly conservative gating may also remove useful memory evidence, limiting the improvement on utility-oriented tasks.

A moderate value provides a better balance. In our main setting, $\lambda=0.1$ keeps the trustworthy metrics low while improving LoCoMo and PreFEval over the OpenClaw baseline. When $\lambda$ further increases, the gate becomes closer to an identity mapping and retains more memories, including noisy or harmful ones. This leads to higher cross-domain FR, sycophancy FR, tool-call FR, and jailbreak ASR. Meanwhile, utility first improves and then slightly drops, suggesting that excessive memory retention can also dilute useful evidence with irrelevant context.

\begin{figure}
    \centering
    \includegraphics[width=0.95\linewidth]{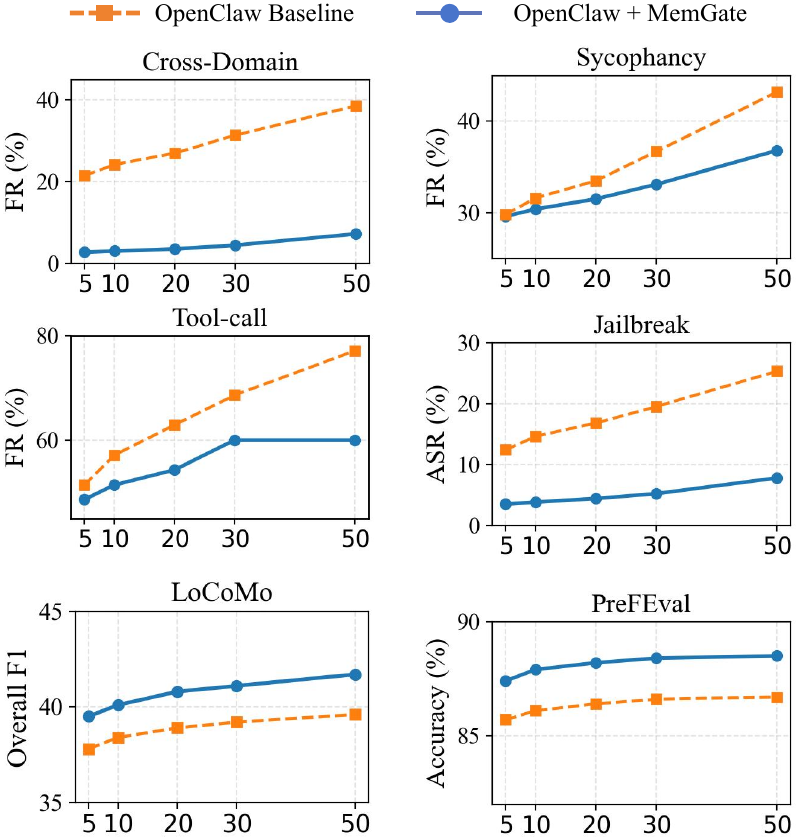}
    \caption{Top-$k$ ablation study using GPT-4o-mini with OpenClaw. Lower values are better for failure rate (FR) and attack success rate (ASR), while higher values are better for Overall F1 and Accuracy.}
    \label{fig:ablation_topk}
\end{figure}

\noindent\textbf{Effect of retrieved memory size top-$k$.}
Figure~\ref{fig:ablation_topk} studies the impact of the number of retrieved memories. Increasing top-$k$ can improve utility because more candidate memories enter the context, making it more likely that the model receives the evidence needed to answer memory-dependent queries. This explains why LoCoMo and PreFEval improve when top-$k$ increases within a moderate range.

However, larger top-$k$ also introduces more irrelevant or unsafe memories. For cross-domain leakage and sycophancy, this means more user profile facts or biased preferences may enter the prompt. For tool-call drift and jailbreak evaluation, larger candidate sets increase the chance that dangerous or misleading memories survive retrieval and influence the final response. Therefore, the trustworthy metrics generally worsen as top-$k$ becomes too large. This result supports using different retrieval budgets for different evaluation goals: a slightly larger top-$k$ is suitable for utility-oriented memory QA, while a smaller top-$k$ is preferable for safety-sensitive settings.

\noindent\textbf{Effect of MemGate encoder.}
We further study whether MemGate is tied to the default \texttt{all-MiniLM-L6-v2} embedding space. Table~\ref{tab:embedding-robustness-size-latency} compares MemGate with four embedding encoders of different dimensionalities, ranging from 384 to 1536. Larger embedding dimensions increase both the input size and the output gate size. 

Overall, MemGate remains effective across all evaluated encoders. Compared with the OpenClaw baseline, every encoder substantially reduces cross-domain leakage, tool-call drift, and jailbreak ASR while improving LoCoMo and PreFEval. Stronger or higher-dimensional encoders bring small additional gains, but the improvement is incremental rather than structural. For example, replacing \texttt{all-MiniLM-L6-v2} with \texttt{text-embedding-3-small} further reduces cross-domain FR from $3.5\%$ to $3.0\%$ and improves PreFEval from $88.2$ to $89.0$, but also increases the MemGate checkpoint size from $35.1$MB to $67.1$MB and fitted retrieval latency from $1.59$s to $1.74$s. This suggests that the main benefit comes from query-conditioned memory gating rather than a specific embedding model.

\section{Conclusion}
   \def\sectionfolder{sections/}%
   Long-term memory is becoming a core component of personal AI agents, our study shows that similarity-based memory retrieval can turn persistent personalization into a trustworthiness risk. Memories that are semantically close to a query may still be contextually inappropriate, causing cross-domain leakage, sycophancy, tool-call drift, and memory-induced jailbreaks. These failures arise because current agents often treat memory as a trusted user context without sufficiently deciding whether the memory should influence the current task.

We propose MemGate, a lightweight query-conditioned retrieval gate that re-ranks candidate memories before they enter the LLM context. By applying a task-conditioned mask to memory embeddings and training the gate with preference-aligned objectives, MemGate provides a modular way to improve contextual isolation without modifying the backbone model or invoking additional LLM calls. MemGate substantially reduces memory-induced failures while preserving, and in several cases improving, long-term memory utility. These results suggest that trustworthy personalization requires agents to not only remember user context, but also determine when memory is legitimate evidence for the current task.%

\bibliographystyle{unsrt}
\bibliography{ref}

\appendices
   \def\sectionfolder{sections/}%
   \section{Proof of Theorem~\ref{thm:bounded_score_perturbation}}
\label{app:proof_of_bounded_score_perturbation}

We provide the proof of Theorem~\ref{thm:bounded_score_perturbation} to
formalize why the positive-mask preservation regularizer stabilizes useful
memory representations. Since MemGate applies an element-wise mask to the memory
embedding, the resulting score change can be bounded by first controlling the
embedding perturbation and then relating it to the change in cosine similarity.
This establishes that a smaller positive mask deficit leads to a smaller
similarity drift for aligned memories.

\noindent \textit{Proof.}
Let $v'_{m^+}=v_{m^+}\odot g^+$. The representation change caused by the gate can be written as
\[
    v_{m^+}-v'_{m^+}
    =
    v_{m^+}\odot(\mathbf{1}_d-g^+).
\]
Thus, the magnitude of the perturbation is bounded by:
\[
\begin{aligned}
    \|v_{m^+}-v'_{m^+}\|_2
    &=
    \|v_{m^+}\odot(\mathbf{1}_d-g^+)\|_2  \\
    &\le
    \|v_{m^+}\|_\infty
    \|\mathbf{1}_d-g^+\|_1 .
\end{aligned}
\]

We next relate this representation change to the change in cosine similarity. For a fixed $q\neq 0$, define the cosine function $f(v)=\frac{q\cdot v}{\|q\|_2\|v\|_2}$. For any $v\neq 0$, its gradient with respect to $v$ satisfies $\|\nabla f(v)\|_2 \le \frac{1}{\|v\|_2}.$ Consider the line segment connecting $v'_{m^+}$ and $v_{m^+}$:
\[
    v_t = v'_{m^+}+t(v_{m^+}-v'_{m^+}), \quad t\in[0,1].
\]
Since $g^+\in[0,1]^d$, each coordinate of $v_t$ can be expressed as:
\[
    v_t = v_{m^+}\odot\left(g^++t(\mathbf{1}_d-g^+)\right),
\]
where the scaling coefficient is no smaller than $g^+$ coordinate-wise. Therefore, $\|v_t\|_2 \ge \|v'_{m^+}\|_2 \ge \gamma .$ Hence, $f$ is $\frac{1}{\gamma}$-Lipschitz continuous along this segment. By applying the mean value theorem, we obtain:
\[
\begin{aligned}
    &\left|S_{\mathrm{MemGate}}(q,v_{m^+})-S(q,v_{m^+})\right|
    =
    |f(v'_{m^+})-f(v_{m^+})| \\
    & \quad \quad \quad \quad \le
    \frac{1}{\gamma}\|v_{m^+}-v'_{m^+}\|_2 
    \le
    \frac{\|v_{m^+}\|_\infty}{\gamma}
    \|\mathbf{1}_d-g^+\|_1 .
\end{aligned}
\]
This concludes the proof. Since our training objective explicitly penalizes the positive mask deficit, the regularizer mathematically limits the similarity drift of aligned memories, successfully discouraging MemGate from inadvertently suppressing useful memory evidence.

\section{Proof of Theorem~\ref{thm:unsafe_suppression}}
\label{app:proof_of_suppression_lower_bound}

\noindent \textit{Proof.} At the ungated reference initialization where $g^+=g^-=\mathbf{1}_d$, the gated cosine similarities identically match the ungated reference scores. Consequently, the relative margin inside the DPO objective evaluates to zero, yielding an initial loss of:
$$\ell_z(\mathbf{1}_d,\mathbf{1}_d)=-\log(\sigma(0))=\log 2.$$

By the definition of $L$-smoothness, for any two vectors $u$ and $v$ within the convex domain, the function $\ell_z$ is upper-bounded by its quadratic expansion:
$$\ell_z(v)\le\ell_z(u)+\langle\nabla\ell_z(u),v-u\rangle+\frac{L}{2}\|v-u\|_2^2.$$

Let $u=(g^+,g^-)$ and $v=(\mathbf{1}_d,\mathbf{1}_d)$. By the theorem's stationarity hypothesis, the gradient at the converged point is zero: $\nabla\ell_z(g^+,g^-)=\mathbf{0}$. Substituting these points into the smoothness condition yields:
$$\ell_z(\mathbf{1}_d,\mathbf{1}_d)\le\ell_z(g^+,g^-)+\frac{L}{2}(\|\mathbf{1}_d-g^+\|_2^2+\|\mathbf{1}_d-g^-\|_2^2).$$

Applying the established upper bounds $\ell_z(\mathbf{1}_d,\mathbf{1}_d)=\log 2$ and $\ell_z(g^+,g^-)\le\epsilon$, we obtain the following inequality:
$$\log 2\le\epsilon+\frac{L}{2}(\|\mathbf{1}_d-g^+\|_2^2+\|\mathbf{1}_d-g^-\|_2^2).$$

Rearranging this inequality provides the analytical lower bound on the total representation perturbation caused by the gating mechanism:
$$\|\mathbf{1}_d-g^+\|_2^2+\|\mathbf{1}_d-g^-\|_2^2\ge\frac{2(\log 2-\epsilon)}{L}.$$

In the specialized scenario where the positive-memory preservation penalty perfectly anchors the aligned memory ($g^+=\mathbf{1}_d$), the subtractive term vanishes. This immediately yields the final corollary bound for unsafe-memory suppression:
$$\|\mathbf{1}_d-g^-\|_1\ge\|\mathbf{1}_d-g^-\|_2\ge\sqrt{\frac{2(\log 2-\epsilon)}{L}}.$$
This concludes the proof.

\section{Detailed Experiments Setups}

\subsection{Utility Benchmark Details}
\label{app:utility_benchmarks}

We evaluate utility from two complementary perspectives. The first group focuses
on general long-term memory capability, measuring whether an agent can retrieve,
maintain, and reason over information across long interaction histories. The
second group focuses more specifically on personalization utility, measuring
whether the system can correctly use beneficial user memories, preferences, or
persona information when such memory is relevant to the current query.

\noindent\textbf{General long-term memory benchmarks.}
These benchmarks evaluate whether a memory-enabled agent can preserve and use
information over extended or incremental interactions.

\begin{itemize}
    \item \textbf{LoCoMo.}
    LoCoMo~\cite{maharana2024evaluating} evaluates very long-term conversational
    memory over multi-session dialogues. It tests whether a model can recover
    and use information from extended interaction histories through tasks such as
    long-term question answering, event summarization, and dialogue generation.
    We report the overall F1 score.

    \item \textbf{MemoryAgentBench.}
    MemoryAgentBench~\cite{hu2026evaluating} evaluates memory agents under
    incremental multi-turn interactions. It covers abilities such as accurate
    retrieval, long-range understanding, and selective forgetting. We report the
    overall accuracy.
\end{itemize}

\noindent\textbf{Personalization benchmarks.}
In contrast to the above benchmarks, the following datasets emphasize whether
the model can use user-specific memories when they are genuinely helpful. These
benchmarks are therefore closer to the intended utility side of personalized
memory systems: relevant memories should improve the response, while failure to
retrieve or apply them reduces personalization quality.

\begin{itemize}
    \item \textbf{PreFEval.}
    PreFEval~\cite{zhao2025llms} measures whether LLMs can infer, memorize, and
    follow user preferences in long-context conversational settings. It contains
    manually curated preference-query pairs across diverse topics and evaluates
    whether the model adapts responses according to previously expressed user
    preferences. We report accuracy.

    \item \textbf{PersonaMem.}
    PersonaMem~\cite{jiang2025know} evaluates dynamic user profiling and
    personalized response selection across multi-session user--assistant
    interactions. Each instance contains evolving user profiles and interaction
    histories, requiring the model to select the response that best matches the
    user's current persona and preferences. We report accuracy.

    \item \textbf{PersistBench beneficial subset.}
    PersistBench~\cite{persistbench} contains both trustworthiness-oriented and
    utility-oriented memory tasks. For utility evaluation, we use its
    beneficial-memory subset, where long-term memory is genuinely relevant and
    should be used. We report beneficial failure rate (Beneficial FR), where a
    failure indicates that the system fails to retrieve or apply the necessary
    memory.
\end{itemize}%

\end{document}